\documentclass[10pt,twocolumn,letterpaper]{article}

\usepackage[pagenumbers]{cvpr} 

\usepackage{graphicx}
\usepackage{amsmath}
\usepackage{amssymb}
\usepackage{booktabs}
\usepackage{url}

\usepackage{utfsym}
\usepackage{pifont}
\usepackage{soul} 
\usepackage{dsfont}

\usepackage[pagebackref,breaklinks,colorlinks]{hyperref}

\usepackage[capitalize]{cleveref}
\crefname{section}{Sec.}{Secs.}
\Crefname{section}{Section}{Sections}
\Crefname{table}{Table}{Tables}
\crefname{table}{Tab.}{Tabs.}

\newcommand\blfootnote[1]{%
  \begingroup
  \renewcommand\thefootnote{}\footnote{#1}%
  \addtocounter{footnote}{-1}%
  \endgroup
}

\makeatletter
\newcommand{\thickhline}{%
	\noalign {\ifnum 0=`}\fi \hrule height 1pt
	\futurelet \reserved@a \@xhline
}
\makeatother

\def\cvprPaperID{*****} 
\def\confName{CVPR}
\def\confYear{2023}

\begin{document}

\title{The 1st-place Solution for CVPR 2023 OpenLane Topology \\ in  Autonomous Driving Challenge}  

\author{
Dongming Wu$^{1\ddagger}$,
Fan Jia$^{2}$,
Jiahao Chang$^{3\ddagger}$,
Zhuoling Li$^{4\ddagger}$,
Jianjian Sun$^{2}$,
Chunrui Han$^{2}$, \\
Shuailin Li$^{2}$,
Yingfei Liu$^{2}$,
Zheng Ge$^{2}$,
Tiancai Wang$^{2}$\\
$^1$ Beijing Institute of Technology,
$^2$ MEGVII Technology, \\
$^3$ University of Science and Technology of China,
$^4$ The University of Hong Kong\\
{\tt\small wudongming97@gmail.com,}
{\tt\small  \{jiafan, wangtiancai\}@megvii.com}
}

\maketitle
\blfootnote{
%
%
$\ddagger$The work is done during the internship at MEGVII Technology.
}
\begin{abstract}
We present the 1st-place solution of OpenLane Topology in Autonomous Driving Challenge.
Considering that topology reasoning is based on centerline detection and traffic element detection, we develop a multi-stage framework for high performance. Specifically, the centerline is detected by the powerful PETRv2 detector and the popular YOLOv8 is employed to detect the traffic elements.
Further, we design a simple yet effective MLP-based head for topology prediction.
Our method achieves 55\% OLS on the OpenLaneV2 test set, surpassing the 2nd solution by 8 points.

\end{abstract}

\section{Introduction}
\label{sec:intro}

OpenLane Topology~\cite{wang2023openlanev2,li2023toponet} is a new perception and reasoning task for  understanding 3D scene structure in autonomous driving.
Given multi-view images that cover the whole panoramic field of view, it requires analyzing the relationship of perceived entities among traffic elements and centerlines.
It consists of four sub-tasks, including centerline  (also named lane) detection, traffic element detection, lane-lane topology, and lane-traffic topology prediction.

In this work, we present a multi-stage framework, which decouples the base detection and topology prediction tasks.
The decoupling strategy is beneficial to the analysis of topology performance since topology prediction is greatly influenced by the detection performance. 
To this end, our solution employs the state-of-the-art 3D/2D detectors for base detection. 
In specific, we modify the query-based PETRv2 detector~\cite{liu2022petr,liu2022petrv2} for 3D lane detection and employ YOLOv8 for 2D traffic detection.
We also design two independent MLP-based heads for lane-lane and lane-traffic topology prediction.
With simple topology heads, \textbf{our method outperforms the second-place solution by nearly 10\% on Top$_{ll}$ and 3\% on Top$_{lt}$.}

\section{Method}

In this section, we present our model in detail.
We first introduce the base model PETRv2 with several architecture modifications.
Then the improved YOLOv8 using several crucial strategies is presented.
Finally, we describe the MLP-based heads for topology reasoning.

\subsection{Lane Detection}

\begin{figure}[t]
  \centering
   \includegraphics[width=0.98\linewidth]{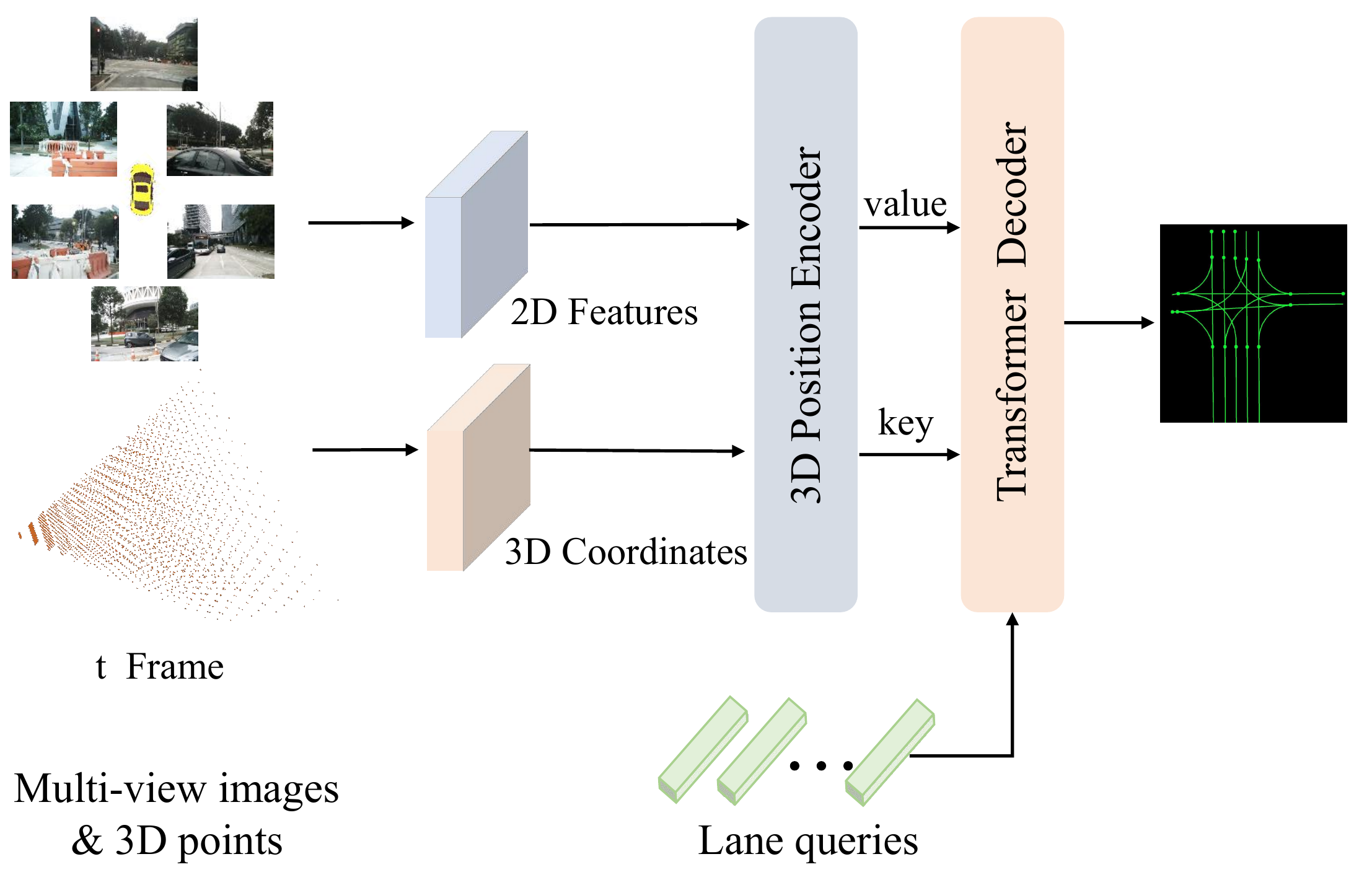}
   \caption{The overall pipeline of lane detection, which builds on PETRv2~\cite{liu2022petrv2} and outputs lane points.}
   \label{fig:petr}
\end{figure}

PETRv2~\cite{liu2022petr,liu2022petrv2} serves as a simple yet strong baseline for multi-view 3D object detection.
It is developed based on DETR framework~\cite{carion2020end,wang2022anchor,wu2023referring} using 3D position embedding.
In specific, it encodes the position information of 3D coordinates into image features, and then uses object query to perceive the generated 3D position-aware features for end-to-end object detection in the transformer decoder.
Following PETRv2, we employ the 3D position-aware features and modify the query representation for centerline detection.
The overall architecture is shown in Fig.~\ref{fig:petr}.

\noindent\textbf{Lane representation.}
Based on the defined lane query, we aim to employ the transformer decoder to predict the 3D control points of Bezier curve.
Specifically, we first randomly initialize $N$ lane queries, \ie,  $Q\! \in \!\mathbb{R}^{N \times 3}$, each of which containing a 3D point. We repeat each point by $M$ times to fit $M$ 3D control points, \ie,  $Q\! \in \!\mathbb{R}^{N \times M \times 3}$.
For each lane query, the $M$ points are flattened and fed into the decoder, outputting the decoded features.
Two independent MLPs are further used to predict the classes and control points, with respect to class head and lane head.
For lane matching and loss calculation, we use Focal loss for the class head and L1 loss for the lane head.
Here, the weight of Focal loss is 1.5 and the weight of L1 loss is 0.0075.

\noindent\textbf{Training strategies.}
As our model has a powerful ability on scaling up, we employ different backbones for improving performance. Besides, we implement several data augmentation methods, including BDA and color gamut augmentation on HSV.

\subsection{Traffic Element Detection}
We utilize YOLOv8\footnote{The official codes we adopt in our competition solution are available at \href{https://github.com/ultralytics/ultralytics}{https://github.com/ultralytics/ultralytics}.} as the 2D detector, which only takes the front image as input and predicts a set of 2D boxes. Based on the characteristics of the OpenlaneV2 dataset, we propose a series of strategies to improve the performance of traffic element detection.

\begin{figure}[t]
  \centering
   \includegraphics[width=0.8\linewidth]{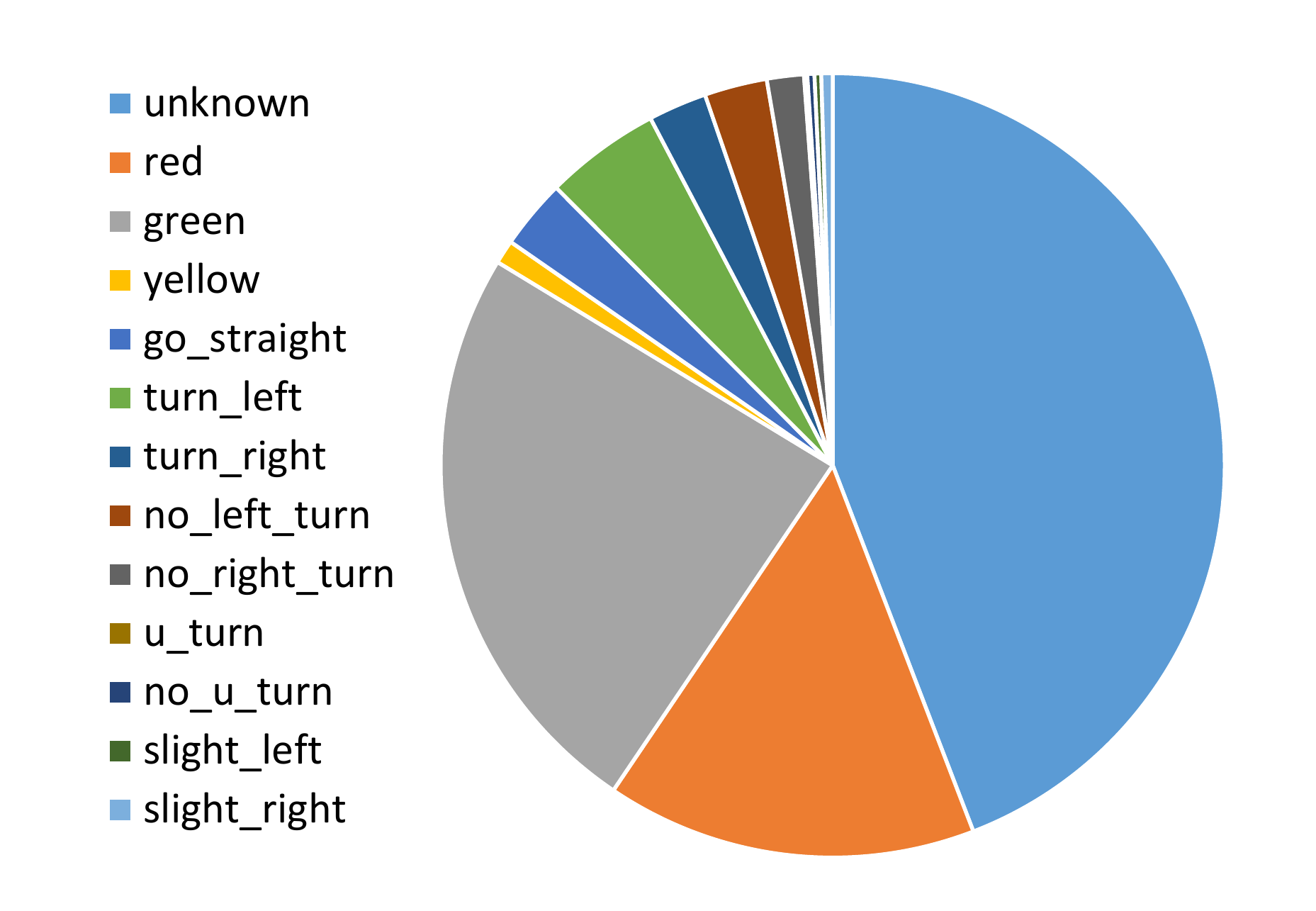}

   \caption{Category distribution on OpenLane-V2 training set.}
   \label{fig:1}
\end{figure}

\noindent\textbf{Strong augmentation.}
Many frames in the OpenLane-V2 dataset share similar scenes and lack of foreground samples, so the detector is easy to be overfitting. We utilize strong data augmentation following YOLOX~\cite{yolox2021}, including mixup, mosaic augment and color gamut augmentation on HSV. It should be introduced carefully since color augmentation may obscure the color of traffic lights, while horizontal flipping can disrupt the direction of traffic signs. 

\noindent\textbf{Reweighting classification loss.}
After visualizing the prediction results, we found that the main difficulty lies in the high similarity of traffic signs, rather than the localization of small-size objects, such as the attribute of  ``turn left'', ``no turn left'' and ``slight left''. Meanwhile, traffic signs have relatively fixed rectangular shapes. Therefore, we choose to reweight the classification loss of these difficult samples only for foreground.

\noindent\textbf{Resampling difficult samples.}
The distribution of categories in the dataset is also worth noting. Through statistics, we observe that the number of ``unknown" in traffic lights accounts for almost half of all annotations, while the number of yellow lights is significantly less than that of green and red lights. The number of the leftover nine traffic signs is only 20\% of the total annotations, as shown in Fig. \ref{fig:1}. Similar to CBGS~\cite{zhu2019class}, we resample the frames in the dataset based on the above category statistics.

\begin{figure}[t]
  \centering
   \includegraphics[width=1.\linewidth]{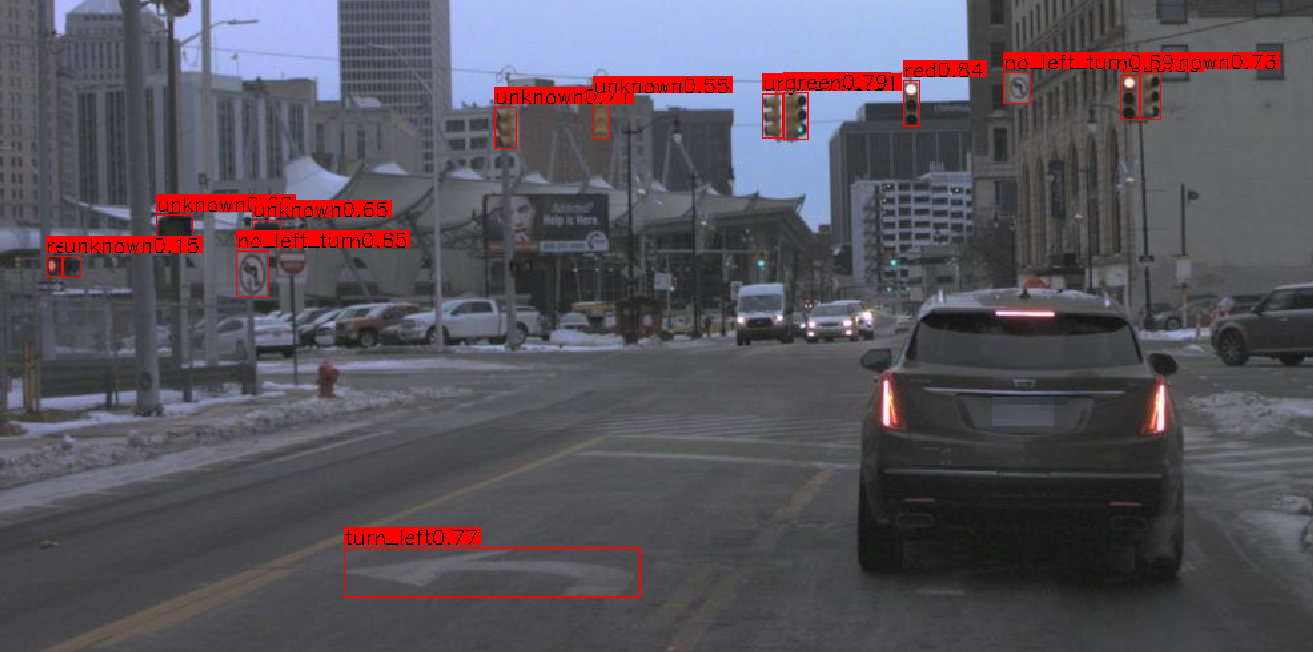}
   \caption{The prediction results from our model on the validation set, annotated with bounding boxes, categories, and confidence.}
   \label{fig:2}
\end{figure}

\noindent\textbf{Pseudo label learning.}
We find that when the ego vehicle forward, the distant traffic elements that first appear in the video frame are usually not annotated due to their small size. Besides, there inevitably be missing annotations in both training and val sets of this dataset, which confuses the model training and leads to sub-optimal performance.
The visualization in Fig.~\ref{fig:2} shows the inference results on the validation set, where the model is only trained on the training set. It can be seen that the model has a high recall and produces an acceptable prediction for those distant small-size objects. We think the high performance detector can be used for pseudo labeling to assist the model in further training. In the ablation section, we find that using the pseudo labels on validation set significantly improves the results compared to those only on the training set. 

\noindent\textbf{Test-time augmentation.}
Test-Time Augmentation can steadily bring benefits during inference phase. 
We only adopt multi-scale testing, as complex transformations may lead to performance degradation. The scale range we adopt is selected between 0.7$ \sim $1.4. The enlarged images can improve the recall of the detector for small-size objects, while the shrunk images are helpful for detecting large-size road signs on the ground in front of the ego vehicle.



\subsection{Lane-Lane Topology}

We collect the decoded features and the predicted lane coordinates from the last decoder layer.
The lane coordinates are transformed into the same dimension as the decoded features using an MLP, and we sum up both two kinds of features.
The summed features are concatenated as the topology size of $N\!\times \! N \! \times \! 2C$, where $C$ represents the feature dimension.
Another MLP further transforms the topology features into a binary topology representation.
We apply Focal loss to supervise the topology learning.

\subsection{Lane-Traffic Topology}
Our lane-traffic topology directly uses the traffic predictions from YOLOv8 as input. The coordinates, category, and confidence score of each box are first concatenated and then projected to $C$-dimensional feature vectors.
Here, we denote the size of traffic features as $T\!\times\! C$.
Therefore, the lane-traffic topology features can be formulated as $N\!\times \! T\! \times\! C$.
With an MLP and sigmoid function, the predicted lane-traffic topology representation is also supervised by Focal loss, similar to the lane-lane topology.

\textbf{
\begin{table}[t]
	\centering
	\small
	\resizebox{0.48\textwidth}{!}{
		\setlength\tabcolsep{12pt}
		\renewcommand\arraystretch{1.0}
		\begin{tabular}{c||c|c|c}
			\hline\thickhline
			Method& Backbone & \#Epoch &DET$_l$ (\%) \\ 
			\hline
            \hline
            Ours &ResNet50 &20 & 18.15\\
            Ours &VOV &20 & 21.01\\
            Ours & ViT-L &20 & 28.11\\
            Ours & ViT-L &48 & 34.16\\
			\hline \thickhline
	\end{tabular} }
	\vspace{-6pt}
	\caption{Ablation study of different backbones and training epochs over  OpenLaneV2 validation set, in terms of DET$_l$ (\%).}
	\vspace{-2pt}
	\label{table:lane_backbone}
\end{table}}

\textbf{
\begin{table}[t]
	\centering
	\small
	\resizebox{0.48\textwidth}{!}{
		\setlength\tabcolsep{13pt}
		\renewcommand\arraystretch{1.0}
		\begin{tabular}{c||c|c|c}
			\hline\thickhline
			Method& Backbone & BDA &DET$_l$ (\%) \\ 
			\hline
            \hline
            Ours & ViT-L & & 34.16\\
            Ours & ViT-L & $\usym{2713}$& 35.28\\
			\hline \thickhline
	\end{tabular} }
	\vspace{-6pt}
	\caption{{Ablation study of BDA over  OpenLaneV2 validation set, in terms of DET$_l$} (\%).}
	\vspace{-2pt}
	\label{table:lane_bda}
\end{table}}

\section{Experiments}
In this section, we first provide some implementation details. Then we evaluate the aforementioned parts in our method on the OpenLaneV2 validation set. The final results on the challenge will be presented as well.

\subsection{Implementation Details}

For lane detection, the size of all input images is resized to 1550$\times$2048.
Our model is implemented with different backbones, including ResNet50~\cite{he2016deep}, VOV~\cite{lee2019energy}, and ViT-L~\cite{dosovitskiy2020image}.
The entire network is optimized by AdamW optimizer with a learning rate of 2e-4, except for the backbones using 2e-5.
The number of lane queries is set to 300.
The model is trained with 20 epoch if not specialized.

For traffic element detection, we statistically analyze the distribution of annotations in vertical direction of the image and crop them into 896$\times$1550 for training efficiency. The classification loss of foreground is reweighted by 2 and the pseudo label loss is weighted by 1. For categories with a quantity ratio of less than 10\%, we implement resampling based on their specific quantity, with a ratio ranging from 5 to 20 times.
We load COCO-pretrained checkpoint and finetune for 20 epochs as our 2D detector baseline. And others all follow the default settings of YOLOv8-x.

For topology prediction, we freeze the 2D traffic and 3D lane detectors, and only train these MLP networks for 10 epochs. The learning rate is set to 2e-4.

\subsection{Lane Performance}

We test different backbones  on the OpenLane-V2 validation dataset for verifying the scaling up of our model.
As shown in Table~\ref{table:lane_backbone}, it can be seen that stronger backbone brings much better performance.
We also train the model using more training time, \ie, 48 epochs, showing large performance improvement (+6.05\% on DET$_{l}$).
Moreover, we investigate the impact of BDA augmentation in Table~\ref{table:lane_bda}. 
Using bird's eye data augmentation (BDA) brings about 1\% performance gain of DET$_{l}$.

\textbf{
\begin{table}[t]
	\centering
	\small
	\resizebox{0.48\textwidth}{!}{
		\setlength\tabcolsep{6pt}
		\renewcommand\arraystretch{1.0}
		\begin{tabular}{c || c c c |c}
			\hline\thickhline
			Method & \#Epoch &  lr scale  & Filter empty gt & DET$_t$ (\%)\\ 
			\hline
            \hline
            YOLOv8-x & 20 & $\times$1 &$\usym{2613}$ & 79.89\\
            YOLOv8-x & 50 & $\times$1 &$\usym{2613}$ & 78.98\\
            YOLOv8-x & 75 & $\times$1 &$\usym{2613}$ & 78.32\\
            YOLOv8-x & 20 & $\times$0.5 & $\usym{2613}$& 72.45\\
            YOLOv8-x & 20 & $\times$0.1 &$\usym{2613}$ & 68.84\\
            YOLOv8-x & 20 & $\times$1 & $\usym{2713}$ & 75.90\\
			\hline \thickhline
	\end{tabular} }
	\vspace{-6pt}
	\caption{{Ablation study of training strategies on traffic element detection over  OpenLaneV2 validation set}.}
	\vspace{-2pt}
	\label{table:ablation_traffic}
\end{table}}

\begin{figure}[t]
  \centering
   \includegraphics[width=1.\linewidth]{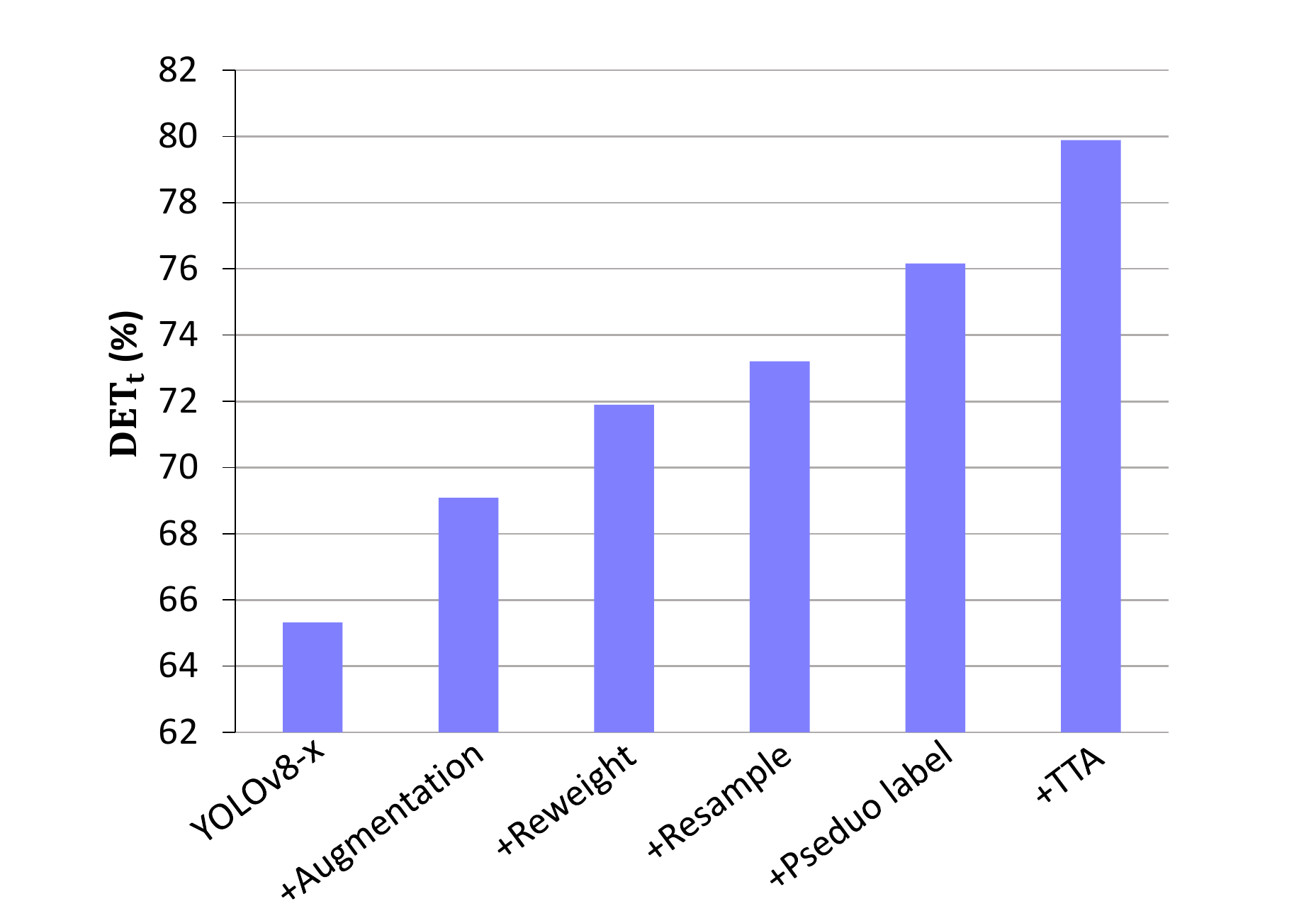}
   \vspace{-12pt}
   \caption{The contribution of each part on traffic element detection over OpenLaneV2 validation set, in terms of DET$_l$ (\%).} 
   \label{fig:4}
\end{figure}

\subsection{Traffic Element Performance}
We evaluate the aforementioned parts in our 2D traffic element detector on the OpenLane-V2 validation dataset in  Fig.~\ref{fig:4}. YOLOv8-x can reach a score of 65.32\% on DET$_l$ without any tricks. Applying strong data augmentation has a performance improvement of 3.77\%.
After adjusting the classification weights, the model performance has improved to 71.90\% from the previous level. Resampling solves the problem of unbalanced sample distribution and brings further improvement, while combining with pseudo labels training brings 1.31\% and 2.95\% improvements, respectively. Finally, we utilize Test-Time Augmentation to improve the performance to 79.89\% without training  models. We also explore other settings during training, as shown in Table~\ref{table:ablation_traffic}. It can be seen that increasing the training epoch from 20 to 75 does not bring any performance improvement. Reducing the learning rate of finutuning stage will seriously damage performance, and samples without ground-truth annotations are still helpful for model training.

\textbf{
\begin{table}[t]
	\centering
	\small
	\resizebox{0.48\textwidth}{!}{
		\setlength\tabcolsep{4pt}
		\renewcommand\arraystretch{1.0}
		\begin{tabular}{cc|ccc}
			\hline\thickhline
		DET$_l$ (\%) & DET$_t$ (\%) & TOP$_{ll}$ (\%) & TOP$_{lt}$ (\%) & OLS (\%) \\ 
			\hline
            \hline
            28.11 & 68.84  &  14.54    & 18.97 & 44.64\\
            28.11 & 79.89	&  14.54	& 21.65	& 48.16	\\
            35.28 & 79.89	&  23.01	& 33.34	& 53.29 \\
			\hline \thickhline
	\end{tabular} }
	\vspace{-6pt}
	\caption{The topology performance (\ie, TOP$_{ll}$ and TOP$_{lt}$) based on different lane  performance DET$_l$ and traffic performance DET$_{t}$  on OpenLaneV2 validation set.}
	\vspace{-2pt}
	\label{table:topo}
\end{table}}

\subsection{Topology Performance}
As topology prediction heavily relies on the performance of lane detection and traffic detection, we conduct ablation experiments with different 2D traffic and 3D lane detection results, as shown in Table~\ref{table:topo}.
The last row uses the best scores of detection (35.28\%	DET$_l$ and 79.89\% DET$_l$), while the scores of other rows (28.11\%	DET$_l$ and 68.84\% DET$_l$) can be seen in Table~\ref{table:lane_backbone} and Table~\ref{table:ablation_traffic} for more details.
It is obvious that the topology performance is improved with higher basic detection performance.

\textbf{
\begin{table}[t]
	\centering
	\small
	\resizebox{0.48\textwidth}{!}{
		\setlength\tabcolsep{4pt}
		\renewcommand\arraystretch{1.0}
		\begin{tabular}{ccccccc}
			\hline\thickhline
		Rank & Name& DET$_l$ & DET$_t$ & TOP$_{ll}$ & TOP$_{lt}$ & OLS \\ 
			\hline
            \hline
            \textbf{1} &\textbf{MFV} &0.36 &	\textbf{0.80} &	\textbf{0.23} &	\textbf{0.33} &	\textbf{0.55}\\
            2 & qcraft2 &0.42 &	0.64 &	0.07 &	0.30 &	0.47\\
            3 & Victory &0.22	& 0.72	& 0.13	& 0.23	& 0.45\\
			\hline \thickhline
	\end{tabular} }
	\vspace{-6pt}
	\caption{{The final leaderboard of OpenLane Topology Challenge}. We ranked 1$^{st}$ on the OpenlaneV2 test set.}
	\vspace{-2pt}
	\label{table:final}
\end{table}}

\subsection{Final Result}

The final result of the OpenLane Topology Challenge is shown in Table~\ref{table:final}.
Our method shows a much stronger performance (+16\%  DET$_t$, +16\% TOP$_{ll}$, +3\% TOP$_{lt}$, and +8\% OLS), compared to the 2$^{nd}$ solution.
Note that the model for final submission is jointly trained on the training and validation sets of OpenlaneV2 but is not ensembled.

{\small
\bibliographystyle{ieee_fullname}
\bibliography{egbib}
}

\end{document}